# Touch Sensing on Semi-Elastic Textiles with Border-Based Sensors

*Samuel Zühlke, Andreas Stöckl and David C. Schedl*

University of Applied Sciences Upper Austria, 4232 Hagenberg im Mühlkreis, Austria

**ABSTRACT**

This study presents a novel approach for touch sensing using semi-elastic textile surfaces that does not require the placement of additional sensors in the sensing area, instead relying on sensors located on the border of the textile. The proposed approach is demonstrated through experiments involving an elastic Jersey fabric and a variety of machine-learning models. The performance of one particular border-based sensor design is evaluated in depth. By using visual markers, the best-performing visual sensor arrangement predicts a single touch point with a mean squared error of 1.36 mm on an area of 125mm by 125mm.We built a textile only prototype that is able to classify touch at three indent levels (0, 15, and 20 mm) with an accuracy of 82.85%. Our results suggest that this approach has potential applications in wearable technology and smart textiles, making it a promising avenue for further exploration in these fields.

## INTRODUCTION

The field of wearable technology and smart textiles has seen rapid growth and development in recent years. A key trend in this field is the use of flexible and tangible surfaces to facilitate user interactions. Traditionally, sensors such as capacitive or resistive sensors are directly placed on the sensing area to detect touch inputs (Mecnika et al. 2015). For instance, capacitive textile touch sensors rely on a 2D matrix of wires to detect touch, which can disrupt the texture, surface structure, and potentially alter the behaviour of the textile (Aigner et al. 2021). Such alterations can compromise the functional qualities of the textile, structural integrity, and aesthetics, limiting the scope of its applications.

To address this issue, we propose a novel approach for touch sensing on semi-elastic textile surfaces without the need to alter or place additional sensors in the sensing area. Our approach involves placing sensors on the border of the textile (cf. Figure 1(a)), leaving the interaction area completely unaltered and free of sensors. The sensors on the border detect stretching caused by interactions in the sensing area, which is measured and used for classification using machine-learning algorithms. Our approach eliminates the need for resistive or capacitive measurements on the textile surface within the touch area, preserving its original texture and surface structure. Furthermore, it allows for a wide range of applications in wearable technology and smart textiles, providing a seamless and unobtrusive way for user interaction. Our research aims to explore the technical challenges involved in developing this border-based approach and evaluate its performance, as well as investigate potential applications and limitations.

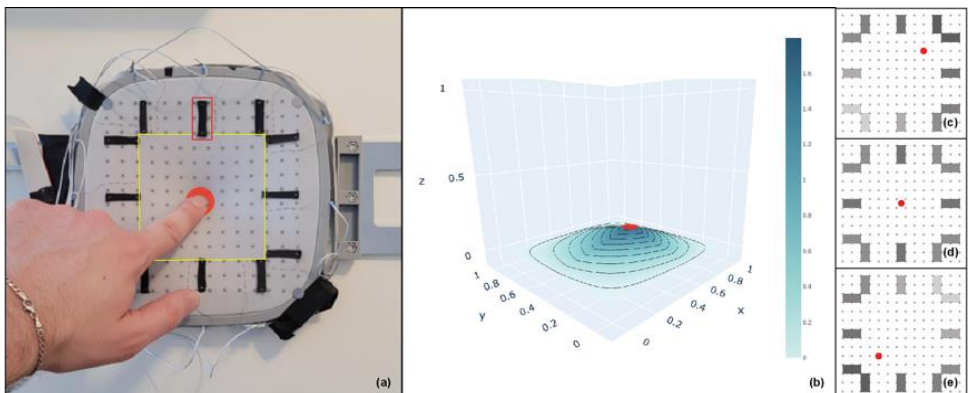

**Figure 1:** Overview of our border-based sensor prototype: A Jersey textile is stretched over a frame and 12 stretch-sensitive patches (a single patch is highlighted in red) are mounted on its border around the touch area (highlighted in yellow) (a). Interactions, like finger presses, lead to a 3D deformation of the fabric as illustrated in (b). By measuring tension at the borders of the fabric we can reconstruct touch points. Different positions and different touch depths lead to varying strain on the border of the sensor, indicated by the brightness of the patch (c-e).

## RELATED WORK

In recent years, there has been significant progress in the development of tactile sensing systems in multiple fields (Chi et al. 2018; Pyo et al. 2021). For example, image sensors have been used to track visual markers within soft synthetic tissue used for robotic grip detection. Together with techniques such as Voronoi segmentation and artificial intelligence, they have been used to improve tactile sensing (Cramphorn et al. 2018; Shimonomura 2019; Yuan et al. 2017).

In the field of robotics and damage detection, electrical resistance tomography (ERT) is used. ERT-based tactile sensors with distributed electrodes can be used in robotic skin to conform to a curved surface (Lee et al. 2021; Park et al. 2020).

Similar studies have explored the potential of electrical impedance tomography (EIT) as a method for soft and stretchable sensor applications, structural damage localisation in composite parts, low-cost and large-area touch sensing using conductive fabric, and its application as a robotic skin (Baltopoulos et al. 2013; Duan et al. 2019; Russo et al. 2017; Silvera-Tawil et al. 2015). In fabric sensing, new algorithms are used to improve the touch localisation accuracy of a knitted or embroidered capacitive and resistive touch sensing system, whereas textile mutual capacitive sensors using resistive and capacitive yarn achieve continuous input of up to three degrees of freedom (Aigner et al. 2021, 2022; Hamdan et al. 2018; Parzer et al. 2018; Pointner et al. 2020, 2022; Vallett et al. 2020).

Utilising the shapeable nature of fabrics, deformable displays together with user defined gestures have been proposed and intelligent robotic manipulation, sensing principles, typical designs, common issues, and applications have been explored (Bacim et al. 2012; Mlakar et al. 2021; Tegin et al. 2005; Troiano et al. 2014).

Overall, a wide range of sensing modalities and technologies can be employed for tactile sensing. While border-based measurement has been explored together with ERT and EIT, and the deformability or stretchability of fabrics has been utilised previously, we believe we are among the first to combine the two paradigms in a novel way. We combined border-based sensing modalities on non-resistive and non-capacitive textiles together with artificial intelligence techniques for accurate and comprehensive tactile sensing systems.

## RESULTS

To implement and validate the proposed border-based approach for touch sensing on semi-elastic textile surfaces, we designed a comprehensive experimental setup. First, to demonstrate the working principle of our sensor, we used a vision-based approach and simulations. We painted a 7 by 7 grid of highly reflective points on the surface of the textile to track its movement and enable reliable and semi-automatic data collection and labelling. In further prototypes, the density of points was increased, and a grid of 14 by 14 points was used. Additionally, we utilised a customised CNC milling machine to create indentations at touch points with predefined depths and random locations on the 80 by 80 mm sensor area. The initial step in our study involved an analysis of motion capture data gathered from the fabric, which we utilised to construct a preliminary digital model. We observed that the movement of the fabric closely resembled a linear surface in three dimensions, leading us to develop a mathematical model that could simulate measurements. We compare our simulated points (from our linear model) to tracked real-world data points over four experiment runs with a total of 2000 frames and found that the overall error across the entire surface is 1.7%, measured as Root Mean Squared (RMS) error. The error between simulation and measurements varies from 0,5% to 2% across the surface as displayed in Figure 2(a).

**Single Indent Localisation**
The recorded real-world data together with simulations were the bases for an assessment of different border-based sensor arrangements for single-indent localisation (as required for single-touch interactions). Therefore, we reconstructed the location of indentations on the fabric by only measuring the stretch between two surface points (i.e., the sensor). The optimal number of sensors and their placement on the surface was the subject of subsequent experiments.

To determine the most optimal sensor configuration, several manually defined arrangements were tested, as illustrated in Figure 2(c). The resulting stretch values for each sensor, for both the physical and mathematical models, were then used as input features for machine learning to reconstruct the indent location. To assess performance, elementary machine-learning models such as random forest, linear regression, and polynomial regression were employed. The data was split randomly into training and test sets at a ratio of 66.6 to 33.3. The graph in Figure 2(b) shows the performance of several configurations with a random forest model for simulated and real-world measurements in Mean Absolute Error (MAE).

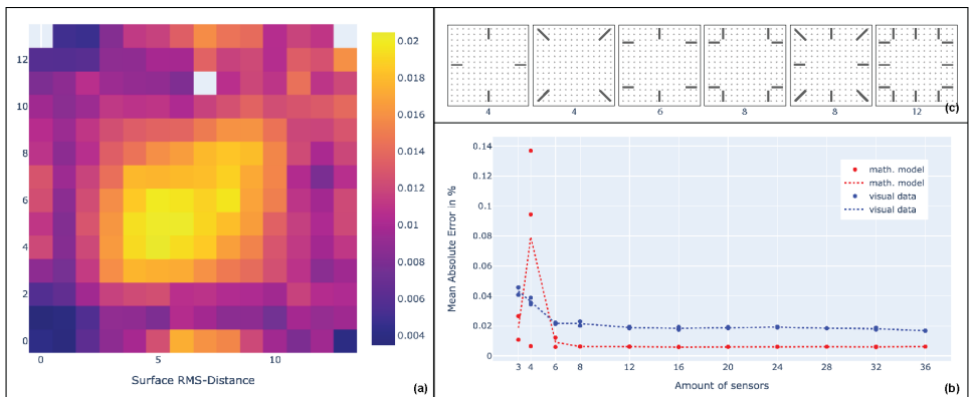

**Figure 2:** Comparison of the mathematical model to real-world measurements: The RMS of the mathematical surface model in comparison to the measured one across the sensor area (a). Note that the white pixels indicate missing optical tracking information. The Mean Absolute Error from the Random Forest sensor pattern evaluation, in relation to the entire touch area, is presented in (b). Various sensor configurations are evaluated, ranging from 3-36 sensors in 26 different arrangements (6 are shown) (c).

The experimental results allowed us to assess varying sensor configurations and machine-learning models for accurately reconstructing touch inputs from border-based sensors. Note that the high MAE visible in the mathematical model with four sensors, in Figure 2(b), is due to the utilised sensor arrangement. Hereby the sensors are arranged in a cross in the centre of the area where the average RMS-Distance for the mathematical model is the highest as can be seen in Figure 2(a). Due to the concentration of sensors in one area, coinciding with the area of the greatest distance between visual and mathematical data, the mathematical model performs worse in this sole instance.

While arrangements with four or less sensors were found to be imprecise, regardless of their placement, the use of six or more sensors yielded better results. A total of 26 different arrangements, ranging from 3 to 36 sensors, were tested and evaluated. In terms of practical production considerations, using fewer sensors minimises the disruption to the fabric structure, and therefore, we selected the best-performing 12-sensor configuration (mean squared error of 1.36 mm) for our textile-sensor prototype.

**Textile-Sensor Prototype**
Based on the findings from the previous experiments (i.e., simulations and optical measurements) we mounted 12 textile sensors on the border of our prototype (cf. Figure 1(a) and Figure 3(a)). The sensors are rectangular patches of conductive fabric that change resistance when stretched. The change in resistance during experiments (random touch points on the sensor area) was recorded and used as input into a machine-learning model. Several models were evaluated and the best-working model for textile, mathematical and visual implementation is a simple Multi-Layer-Perceptron.

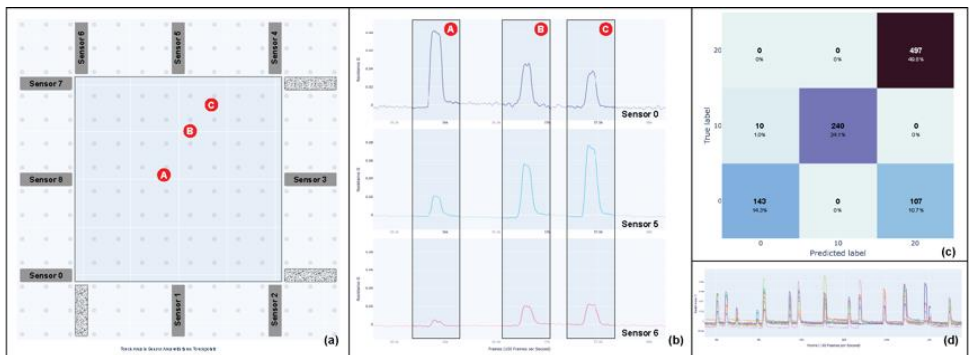

**Figure 3:** Here the sensor arrangement in the sensor area is illustrated, along with the touch area and three indentations (a). Additionally, the time plots for sensor data from sensors 0, 5, and 6 at the duration of the indentations are displayed in (b). The performance of the MLP in predicting the indent levels, as compared to the actual indent levels of a test matrix, is presented in the form of a confusion matrix (c). The raw sensor data stream as it is received from the sensors is depicted in (d). Note that at the time of measurement 3 of the 12 sensors were defective and recorded noise data. Thus, they were excluded from the experiments and graphs.

The training was performed with a set of 997 touch points and the model was evaluated on 499 test points. The accuracy of touch classification at three different indent levels (0, 15, and 20 mm) from any given touch event using sensor data is 82.85%. In comparison, the simulated sensor achieves a validation accuracy of 91.17% for classifying the indentation depth (0, 15, or 20 mm).

**METHOD**

In developing our sensor prototype, we selected a knit jersey fabric as the primary material that was stretched over a rectangular frame with a sensing area of 125 by 125 mm. The textile sensors themselves were created from rectangular pieces of EeonTex™ Conductive Stretchable Fabric which were cut and sewn onto the jersey using highly conductive, polyamide silver plated yarn from Madeira. To facilitate repeated and precise touch events, an industrial embroidery frame was utilised, which enabled the prototype to be mounted onto a customised CNC milling machine with a stylus-based touch point attachment for testing purposes. All sensors share a common ground and are individually connected to a measurement unit. To extract measurement data, the hardware platform prototyping kit CY8CPROTO-063-BLE from Infineon is used, with the data subsequently being recorded in a CSV file for further analysis. This setup allowed us to measure and validate the performance of our proposed approach for touch sensing on semi-elastic textile surfaces.

For tracking the surface, a set of six motion capture cameras were used to capture the textile behaviour in three dimensions, allowing us to measure the stretching of the sensor's border caused by touch inputs in the sensing area at a rate of 100 frames per second. We used Flex-3 cameras from Optitrack and analysed the tracking data with Optitrack's Motive software in version 2.2. To enable the accurate representation of the fabric's movement during indentation and its behaviour in response to touch, we transformed the 3D coordinates into a scaled coordinate system, resulting in the rotated, scaled, and translated points being represented within a zero-to-one range on each axis. The position of reflective points compared to the resting state, enables the calculation of surface parameters and the amount of stretch between measured coordinates. Additionally, the CNC milling machine's head was also tracked optically for precise measurements in the same coordinate system as the surface.

The stretching properties of both the physical fabric and a corresponding mathematical surface model were measured at various locations, including at potential sensor locations. An artificial sensor was placed between two selected points on a grid, and the degree of stretch was recorded. To determine the most optimal sensor configuration, several different arrangements were tested, as illustrated in Figure 2(c). The resulting stretch values for each sensor, for both the physical and mathematical models, were then used as features in a machine-learning approach. To assess the performance, elementary machine-learning models such as random forest, linear regression, and polynomial regression were employed. Alongside, different standard models from TensorFlow (Martín Abadi et al. 2015), Skicit-learn (Pedregosa et al. 2011) and PyTorch (Paszke et al. 2019) were tested. Ultimately a TensorFlow Keras Sequential model was chosen. The MLP model used in the code has two hidden layers, each with 64 neurons and ReLU activation function, and an output layer with 4 neurons and SoftMax activation function, resulting in a total of 196 neurons.

## CONCLUSION

We propose a border-based approach for touch sensing on semi-elastic textile surfaces. By placing sensors on the border of the textile, we leave the interaction area completely unaltered and free of sensors. The sensors on the border detect stretching caused by interactions in the sensing area, which is then measured and classified using machine-learning algorithms.

In our experiments, we show that a simple linear surface model is precise enough for designing optimal sensor configurations and verify this experimentally with simulations and optical tracking. For our experiments, we utilised an elastic Jersey fabric stretched over a rectangular frame with a sensing area of 125 by 125 mm. For this setup, we found an optimal sensor arrangement with 12 border-based sensors, that is capable of reconstructing touch points with an error of 1.36mm and classifying three indentation levels with an accuracy of 91.17% on the simulated mathematical data and 82.85% on the sensor data.

Furthermore, we built the first border-based-sensing and textile-only prototype that is able to classify the indent of touch points. The lower classification accuracy of our prototype can be attributed to the presence of noise and errors that may have resulted from hand-cutting the sensor patches. Additionally, our physical prototype uses resistance measurements as input to the machine-learning model, while our simulation and tracking experiments employ distance measurements directly. Therefore, the resistance data might introduce an additional level of complexity when compared to our preliminary analysis.

In the future, we want to investigate if the machine-learning models can be further optimised and how our prototype will perform in real-world applications with more complex touch inputs and varying environments. Based on the findings of our study, additional attention will be devoted to the detection of the point of interaction, rather than solely relying on the indent level. Further exploration is necessary to determine whether alternative sensor arrangements or configurations may be advantageous in this regard. Nevertheless, the results of our experiments demonstrate that the proposed approach is effective and might be applied in various wearable technology and smart textile applications in the future. Our approach allows a seamless and unobtrusive way for users to interact with textile surfaces.

## ACKNOWLEDGMENT

The authors would like to acknowledge the valuable contributions of our colleagues Roland Aigner, Andreas Pointner and Thomas Preindl for their technical assistance and collaborative efforts, which were instrumental in achieving our research objectives.